\documentclass[letterpaper]{article}

\usepackage[numbers]{natbib}

\usepackage[utf8]{inputenc} 
\usepackage[T1]{fontenc}    
\usepackage[hidelinks]{hyperref}       
\usepackage{url}            
\usepackage{booktabs}       
\usepackage{amsfonts}       
\usepackage{nicefrac}       
\usepackage{microtype}      
\usepackage{xcolor}         
\usepackage{amsmath,amssymb,amsthm}
\usepackage{bm}
\usepackage{graphicx}
\usepackage{pgfplots}
\usepackage[justification=centering,singlelinecheck=false]{subfig}
\usepackage{pifont}
\usepackage{makecell}
\usepackage{centernot}
\usepackage{authblk}

\newtheorem{dfn}{Definition}

\usepackage{nameref}
\usepackage{ifthen}
\newboolean{REVISE}
\setboolean{REVISE}{false}
\newcommand{\revi}[1]{\ifthenelse{\boolean{REVISE}}{\textcolor{blue}{#1}}{#1}}

\newboolean{REVISEF}
\setboolean{REVISEF}{false}
\newcommand{\revif}[1]{\ifthenelse{\boolean{REVISEF}}{\textcolor{blue}{#1}}{#1}}

\let\paragraph\subsection

\usetikzlibrary{positioning}

\usepackage{enumitem}

\usepackage{pgfplots}
\pgfplotsset{width=10cm,compat=1.9}

\usepgfplotslibrary{external}
\def\beq{\begin{equation}}
\def\eeq{\end{equation}}
\def\bea{\begin{eqnarray}}
\def\eea{\end{eqnarray}}

\newcommand{\X}{\vec{X}}

\renewcommand{\vec}[1]{{\mathbf #1}}

\newcommand{\indep}{\perp\mkern-9.5mu\perp}
\newcommand{\dep}{\centernot{\indep}}

\title{Explainable AI needs formalization}

\author[1,2,3,*]{Stefan Haufe}
\author[2]{Rick Wilming} 
\author[2]{Benedict Clark}
\author[2]{Rustam Zhumagambetov}
\author[2]{Ahc\`{e}ne Boubekki}
\author[2]{Jörg Martin}
\author[2]{Danny Panknin}
\affil[1]{Technische Universit\"at Berlin, Berlin, Germany}
\affil[2]{Physikalisch-Technische Bundesanstalt, Berlin, Germany}
\affil[3]{Charité -- Universitätsmedizin Berlin, Berlin, Germany}
\affil[*]{Corresponding author (\texttt{haufe@tu-berlin.de})}
\begin{document}

\maketitle

\begin{abstract}
\revi{The field of ``explainable artificial intelligence'' (XAI) seemingly addresses the desire that decisions of machine learning systems should be human-understandable. However, in its current state, XAI itself needs scrutiny. Popular methods cannot reliably answer relevant questions about ML models, their training data, or test inputs, because they systematically attribute importance to input features that are independent of the prediction target. This limits the utility of XAI for \revi{diagnosing and correcting data and models, for scientific discovery, and for identifying intervention targets}. The fundamental reason for this is that current XAI methods do not address well-defined problems and are not evaluated against targeted criteria of explanation correctness. Researchers should formally define the problems they intend to solve and design methods accordingly. This will lead to diverse use-case-dependent notions of explanation correctness and objective metrics of explanation performance that can be used to validate XAI algorithms.}
\end{abstract}

\section*{Introduction}

The use of machine learning (ML) holds great promise in many fields, including high-risk domains such as medicine. Regulations like the European AI Act demand that ``high-risk AI systems shall be designed and developed [...] to enable deployers to interpret the system's output and use it appropriately'' \citep{euAIActProposal2021}. This need for ``human-understandable'' descriptions of ML models is seemingly addressed by the field of ``explainable artificial intelligence'' (XAI). \revif{However, this paper shows that the current algorithm-first paradigm of XAI development has led to methods whose results are commonly misinterpreted, and whose ability to serve purposes commonly associated with XAI, such as quality assurance for ML, is thereby compromised. To overcome biased or confounded XAI results, the fields requires formalization that goes beyond current axioms and desiderata, which mostly spare out the most important issue of explanation \emph{correctness}. XAI methods should be developed to solve well-defined problems following a process that i) defines appropriate formal criteria of XAI correctness and fitness for relevant purposes and, ii) employs theory and empirical validation based on ground-truth data to assess a given method's adherence to these criteria. This paper further argues that XAI methods targeted at explaining the model function only are insufficient to address desired downstream purposes such as diagnosing and correcting data and models, scientific discovery, and identifying intervention targets.}

\paragraph*{\revi{Supervised Machine Learning}}
\label{sec:supervisedML}
\revi{Machine learning (ML) is concerned with learning functions from data. The most common paradigm of supervised ML corresponds to finding a \emph{model} function $f_{\bm \theta}$ parameterized by a vector ${\bm \theta}$ such that $\hat{\vec{y}} = f_{\bm \theta}(\vec{x})$ approximates a \emph{target variable} $y$. This function is learned from $n$ pairs of observed \emph{training data} $\{(\vec{x}(k), {y}(k))\}_{k=1}^n$, where $\vec{x}(k) = [x_1(k), \hdots, x_q(k)]^\top, \; 1 \leq k \leq n$ are the model's \emph{inputs} and ${y}(k)$ are the corresponding \emph{outputs} (or \emph{targets}). The $q$ individual dimensions $x_i$ of the potentially high-dimensional inputs $\vec{x}$ are called \emph{features}. We assume the target ${y}$ to be a scalar quantity, corresponding to a regression or classification setting.} 
\revi{The goal of supervised ML is not only to fit the training data but also to make accurate predictions on new \emph{test inputs} $\vec{x}$ for which no corresponding outputs are observed.}

\revi{An example would be a neural network predicting the clinical outcome for patients in critical care from clinical and demographic patient characteristics. Here, different characteristics such as age or the presence of pre-existing conditions correspond to individual input features $x_i$, while the outcome of interest, such as death, corresponds to the target $y$. The neural network $f_{\bm \theta}$ learns the mapping between inputs and targets, where ${\bm \theta}$ represents the learnable parameters of the network.}
\revif{An introduction into supervised ML can be found in \citep{bishop2006pattern}.}

\paragraph*{\revi{Explainable Artificial Intelligence (XAI)}}
\label{sec:XAI}
\revi{``Explainable Artificial Intelligence'' (XAI) is an umbrella term for algorithms aiming to provide insight into the properties of ML models, their training data, a given test input submitted to the model, and/or the interplay between these. 
The predominant XAI paradigm is \emph{feature attribution}, which refers to attributing an ``importance'' score $e_i$ to each input feature $x_i$. 
A distinction is made between \emph{global} methods, where the attribution $\vec{e} = [e_1, \hdots, e_q]^\top$ is a property of the model only, and \emph{local} methods, where the attribution $\vec{e}(\vec{x}) = [e_1(\vec{x}), \hdots, e_q(\vec{x})]^\top$ additionally depends on the input. For the example in described in \nameref{sec:supervisedML}, global methods would assign each predictor, such as age, a constant importance score, whereas local methods would assign a score specific to each patient.}
\revif{Recent reviews of predominant XAI paradigms, approaches, and associated claims can be found in \citep{tjoa2020survey,minh2022explainable}. A comprehensive social science perspective on XAI is added in \citep{miller2019explanation}.}

\section*{\revi{Desired} purposes of XAI}
\label{sec:purposes}
\revi{The popularity of XAI tools, including feature attribution methods, rests on their promise to facilitate one or more of the following purposes:} 
 
\paragraph*{\revi{Model and data diagnostics and correction}}
It is often of interest to know which features of a dataset or of a single sample an AI system ``bases'' its decision on. 
\revi{This information would then be used to judge whether a model performs in unexpected or undesired ways, and whether its training data has unexpected or undesired properties.}

In mammographic data analysis, a radiologist would likely trust a cancer diagnosis made by an AI if told that the decision was based on a patch of tissue they themselves identify as cancerous. Conversely, if the XAI method assigns high ``importance'' to features that are known not to be associated with cancer, this might lead to the dismissal of the model itself as being wrong~\citep{saporta_benchmarking_2022}. 

Ribeiro et al.~\cite{ribeiroWhyShouldTrust2016} state ``A model predicts that a patient has the flu, and LIME highlights the symptoms in the patient’s history that led to the prediction. Sneeze and headache are portrayed as contributing to the ‘flu’ prediction, while ‘no fatigue’ is evidence against it. With these, a doctor can make an informed decision about whether to trust the model’s prediction''.

\revi{In a similar vein, it is expected that XAI can help to verify that protected attributes (e.g., gender, race) do not influence model decisions.}

One may also be interested in whether a model bases its decisions on confounding variables. Confounders induce correlations between training in- and outputs that can be used by the model for prediction. This can be problematic if the same correlations are not present in a testing context, leading the model to perform poorly. 
Lapuschkin et al.~\cite{lapuschkinUnmaskingCleverHans2019} study a case in which a watermark in images indicates such confounding, and use XAI methods in the process of identifying this effect. 

\revi{Anders et al.~\cite{anders2022finding} have proposed adjustments to the models themselves to deal with confounding. Similarly, 
Wang et al.~\cite{wang2021gam} advocate to actively manipulate models that are diagnosed to use undesired features. These examples illustrate the desire to use XAI to guide ML quality control.}


\paragraph*{Scientific discovery} 
Various authors \citep{samek2019towards, jimenezLunaDrugDiscoveryExplainable2020,tideman_automated_2021,watson_interpretable_2022,wong_discovery_2024} argue that XAI methods could be used to discover novel associations between variables, generating new hypotheses that could be tested in future experiments. For example, a disease might be related to a complex interaction of multiple previously unknown genetic factors. Such an interaction might not be amenable to classical statistical analysis, but it could be used by an ML model. The promise of XAI methods is then to identify the features contributing to the interaction. 

\paragraph*{Identification of intervention targets}
XAI is frequently used to identify features, the manipulation of which would change a model's output, a task also known as algorithmic recourse. For example~\citep[see][]{ustun_actionable_2019}, a bank might use an ML model to predict the return probability of a loan. For a known model and a given input, XAI would then be able to recommend changes of input variables (e.g., `increase salary') to turn a negative outcome into a positive one. 
\revif{It is typically assumed that such interventions would not only affect the model output but also the target variable the model was trained to predict (e.g., the probability of default).}
In an intensive care unit, an ML model might be used to predict mortality or other severe outcomes. Using XAI to identify possible intervention targets, such as medications, in this context \citep[e.g.,][]{ates_counterfactual_2021} implies that interventions have real-world consequences on the target variable beyond just changing the model output.

\revi{While the use of XAI to address such purposes is appealing and may seem intuitive, all discussed purposes require information about the data-generating process that, as we outline below, is not provided by current feature attribution methods.}

\section*{\revi{Current XAI does not serve desired purposes}}
\revi{Wilming et al.~\cite{wilming2022scrutinizing,wilming2024gecobench} introduce the statistical association property (SAP) for feature attribution methods, which is defined as follows:}
\revi{
\begin{dfn}[Statistical Association Property, SAP]
\label{dfn:sap}
A feature attribution method $\vec{e}$ possesses the SAP if any (statistically significant) non-zero importance attribution to a univariate feature $x_j$ indicates a statistical association with the target: ``$e_j$ indicates importance'' $\Rightarrow x_j \dep y$.
\end{dfn}}
\noindent In the following, we discuss results presented in
Haufe et al.~\cite{haufe2014interpretation,kindermansLearningHowExplain2017,wilming2023theoretical}, showing that a wide range of popular local and global feature attribution methods in fact do not possess the SAP, thus prohibiting conclusions about associational or even causal relations between features and target on the basis of these methods. 

\revif{In \nameref{sec:implications}, we further argue that the use of feature attributions for all of the above-mentioned explanation purposes amounts to asserting that the SAP holds; in other words, the SAP is a necessary property for XAI methods to serve these purposes. As a consequence, the discussed methods fall short of reliably serving the purposes mentioned in \nameref{sec:purposes}.}

\begin{figure}[htbp]
\centering
\subfloat[$c=0.8$]{
    \includegraphics[scale=0.35]{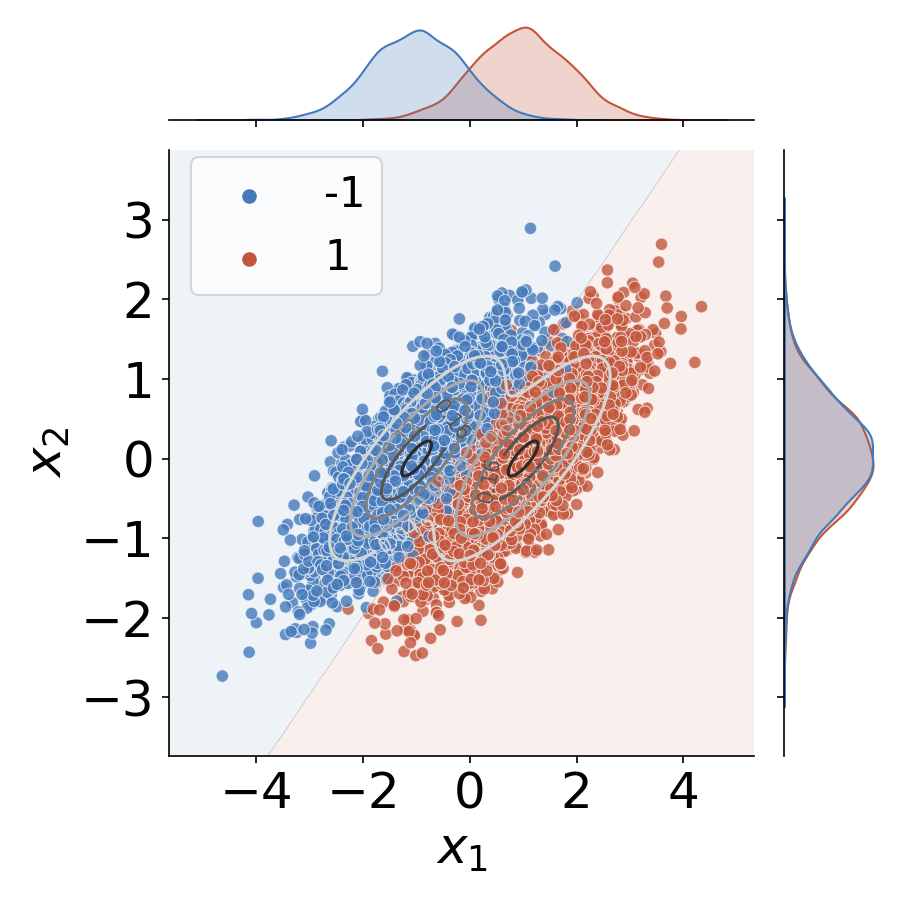}}
\hspace{0.5cm}
\subfloat[$c=0$]{
    \includegraphics[scale=0.35]{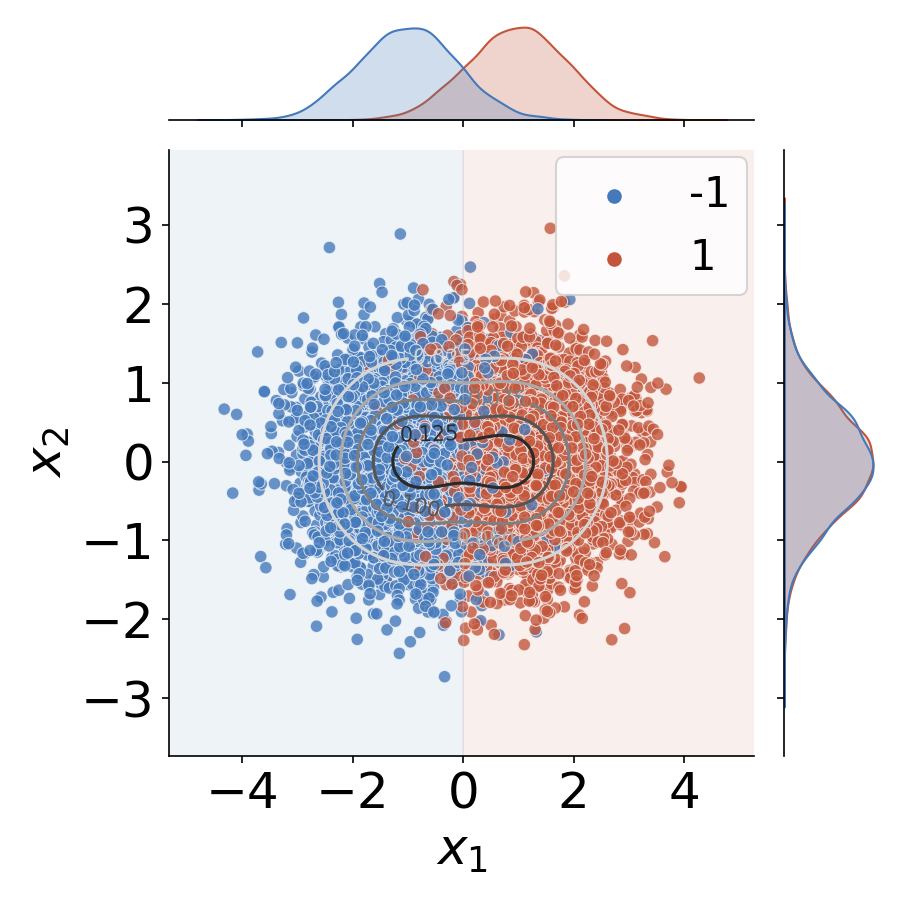}}
%
\\
\subfloat[Suppressor ($X_2$) and Collider ($X_1$)]{
\includegraphics[width=0.35\textwidth]{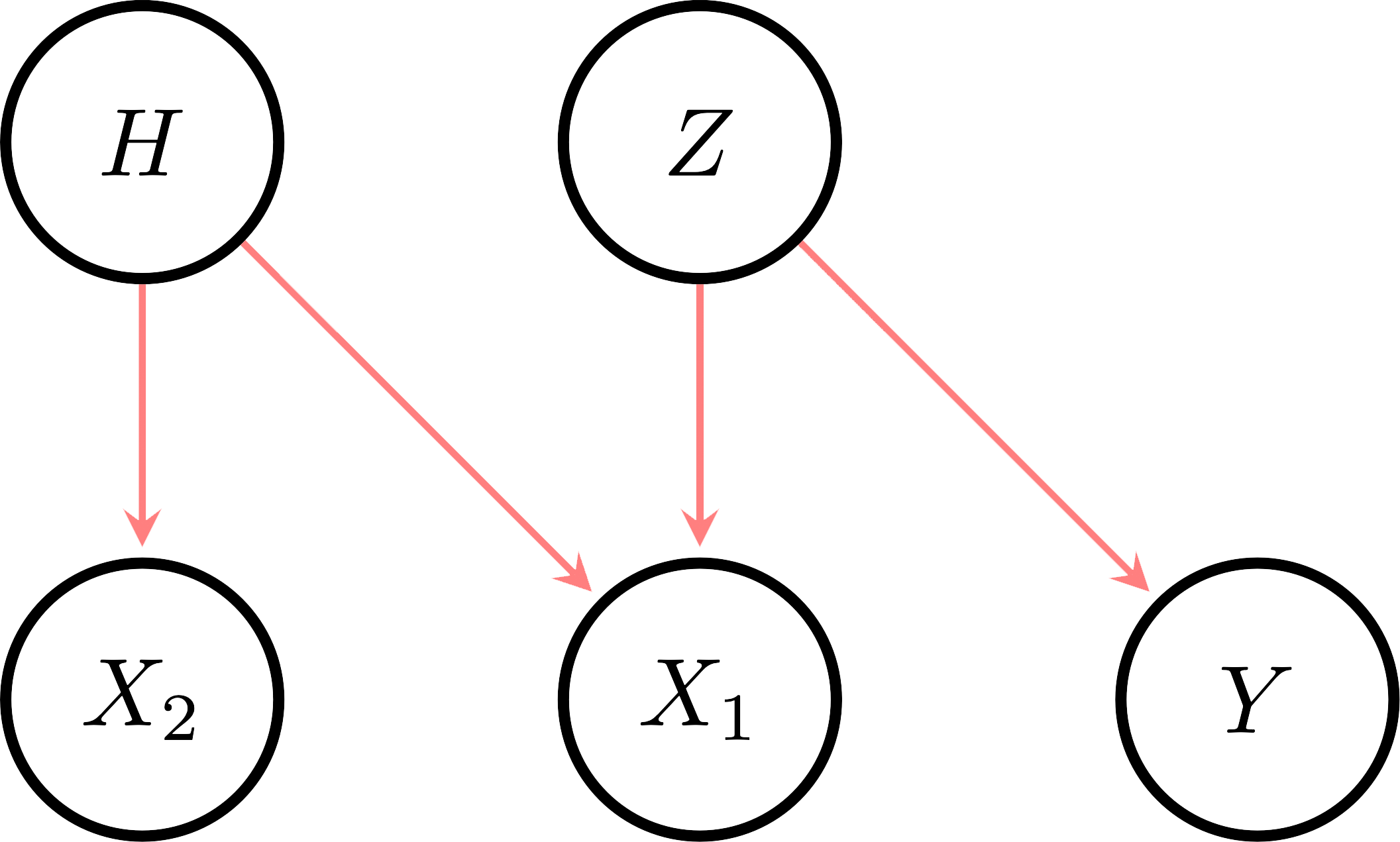}
\hspace{1cm}
\includegraphics[width=0.35\textwidth]{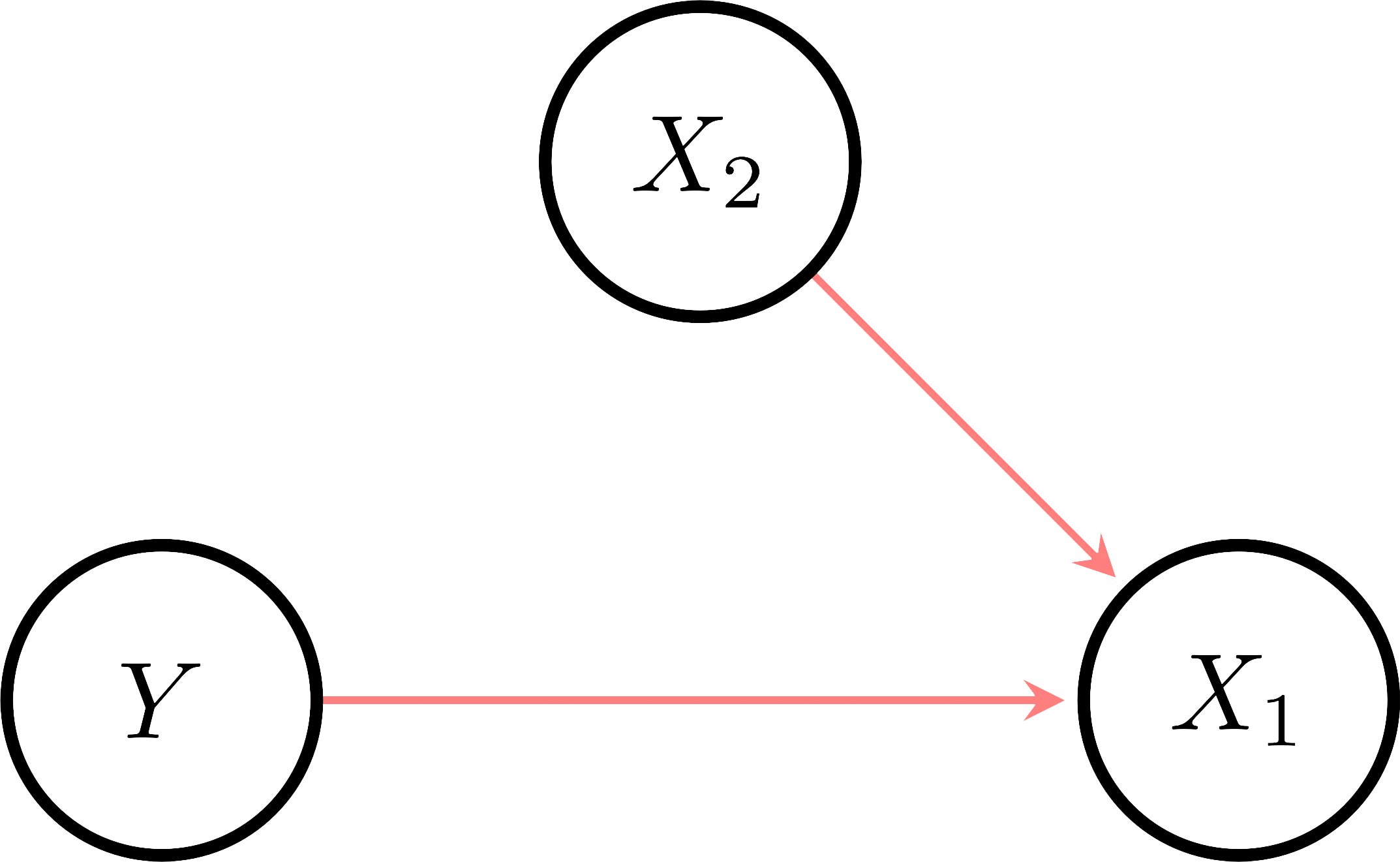}}
\caption{a/b) Data sampled from the generative model (Example A) introduced in \nameref{sec:examples} \citep{wilming2023theoretical} for two different correlations $c$ and constant variances $s_1^2 = 0.8$ and $s_2^2=0.5$. Boundaries of {the} Bayes-optimal decisions are shown as well. The marginal sample distributions illustrate that feature $X_2$ does not carry any class-related information. c) Causal structure of the data in Examples~A (left) and B (right). $X_2$ is a so-called suppressor variable that has no statistical association with the target $Y$, although both influence feature $X_1$, which is called a collider. Figure partially adopted from Wilming et al.~\cite{wilming2023theoretical}.}
\label{fig:fig1}
\end{figure}

\paragraph*{Two minimal examples of classification problems}\label{sec:examples}
In Haufe et al.~\cite{haufe2014interpretation}, the two-dimensional classification problem $\X = \vec{a} Z + \vec{H}, \, Y = Z$ ({Example A}) is introduced, with $\vec{a}=(1, 0)^{\top}$, $Z \sim \text{Rademacher}(1/2)$, and $\vec{H} \sim N(\vec{0}, {\bm \Sigma})$ with covariance 
${\bm \Sigma} = \bigl( \begin{smallmatrix}
s_1^2 & c s_1 s_2 \\ c s_1 s_2 & s_2^2 \;,
\end{smallmatrix} \bigl)$, where $s_1$ and $s_2$ are non-negative standard deviations, and $c \in [-1, 1]$ is a correlation. In this example, only feature $X_1$ is correlated with the classification target $Y=Z$ through $a_1=1$. \revi{By} contrast, $X_2$ is independent of $Y$ since $a_2$ = 0. Both features are correlated through the superposition of additive noise $\vec{H}$ with covariance $\bm \Sigma$. A depiction of data generated under this model is provided in Figure~\ref{fig:fig1}~(a/b). For $c \neq 0$, the Bayes-optimal bivariate linear classification model $f_{\vec{w}, b}(\vec{x}) = \vec{w}^\top \vec{x} + b$ can reduce the contribution of $\vec{H}$ from $X_1$ using information contained in $X_2$, and thereby estimate $y$ as $\hat{y} = f_{\vec{w}, b}(\vec{x})$ more precisely\revi{, compared to} what would be possible using $X_1$ alone
\citep{haufe2014interpretation}. To this end, it needs to put non-zero weight $w_2 = -\alpha c s_1/s_2 $ on $X_2$, where $\alpha = (1+(cs_1/s_2)^2)^{-\frac{1}{2}}$ and $||\vec{w}||_2=1$. This shows that linear models can assign arbitrarily high weights to features, like $X_2$, that have no statistical association with $Y$.\\
\noindent An even simpler example is given by the generative model $X_1 = Y - X_2 $ ({Example B}), where $X_2$ and the target $Y$ are independent \citep{haufe2014interpretation}. Here the Bayes-optimal linear model with weights $w_1 
= w_2 = 1$ completely removes the nuisance term $X_2$ from $X_1$ to recover $Y$, yielding a model output that is statistically independent of $X_2$. Such examples question the notion of a model ``using a feature'' or ``basing its decision on a feature''.

\paragraph*{Suppressor variables}
Features like $X_2$ in Examples~A and B, which improve predictions without being predictive themselves, are called suppressor variables in causal terminology \citep{congerRevisedDefinitionSuppressor1974}. Causal diagrams \citep[see][]{pearl2009causality} of the generative models in both examples are provided in Figure~\ref{fig:fig1}~(c).
Broadly speaking, any variable that is not informative (statistically associated with the target) itself but statistically related to an informative variable (e.g., modulating it through an independent mechanism) is a suppressor. Suppressors occur widely in real-world datasets and hamper model interpretation. 
As one example, the prevalence of a disease may be related to a person's blood pressure but not their age. However, as blood pressure has an age-dependent baseline, the model might need to adjust its prediction with respect to that baseline in order to remove irrelevant variance introduced by age. Age, thereby, becomes a suppressor variable.
In image classification, non-discriminative features such as lighting or weather conditions, or non-discriminative objects occluding class-specific objects, could be suppressors. 

\paragraph*{Existing feature attribution methods attribute importance to variables unrelated to the target}
Recent theoretical and empirical research has shown that various popular feature attribution methods consistently assign importance to suppressor variables \citep{haufe2014interpretation,kindermansLearningHowExplain2017,wilming2022scrutinizing,wilming2023theoretical,clark2024xai}. We will call such methods suppressor attributors from here on.
Kindermans et al.~\cite{kindermansLearningHowExplain2017} showed analytically that the importance scores returned by gradient-based techniques \citep{baehrensHowExplainIndividual2010}, LRP \citep{bachPixelWiseExplanationsNonLinear2015}, and DTD \citep{montavonExplainingNonLinearClassification2017} reduce to the weight vector $\vec{w}$ in case of linear models. Thus, these methods are suppressor attributors. In Wilming et al.~\cite{wilming2023theoretical}, the latter was shown also for Shapley values \citep{shapley1953value} 
and their approximations such as SHAP \citep{lundbergUnifiedApproachInterpreting2017,aasExplainingIndividualPredictions2020}, as well as for LIME \citep{ribeiroWhyShouldTrust2016}, integrated gradients \citep{sundararajanAxiomaticAttributionDeep2017}, and counterfactual explanations \citep{wachterCounterfactualExplanationsOpening2017}. 
A list of some suppressor attributing methods is provided in Table~\ref{tab:tab1}.

\paragraph*{\revif{Implications of suppressor attribution on explanation purposes}}\label{sec:implications}
Since suppressor variables have no statistical or causal association with the target variable, suppressor attributors do not possess the SAP, which has implications regarding their expected utility for the purposes introduced in \nameref{sec:purposes}. {For example, s}uppressor features may often not coincide with prior expectations of an expert. Therefore, suppressor attributors cannot be used in a straightforward way to validate or invalidate \revi{the correctness and fitness-for-purpose of} models or their decisions using expert knowledge as anticipated by Ribeiro et al.~\cite{ribeiroWhyShouldTrust2016}. 
Similarly, high importance on a protected attribute does not necessarily mean that the method ``uses'' this attribute for prediction. The model may also just remove variance related to that attribute from other informative variables.
Moreover, since it cannot be concluded that the highlighted features are part of previously unknown interactions or are causally related to the output, these methods cannot be reliably used to facilitate scientific discoveries.
\revif{While counterfactual and algorithmic recourse techniques are able to point to changes in inputs that have desirable effects on model outputs, such counterfactuals should not be expected to have an equivalent effect on the real-world target themselves. As optimal models may rely on causal effects of the target as well as on suppressor variables, counterfactual explanations may also typically include such variables, the manipulation of which can have no effect on the target in the real world. Recommendations based on such counterfactuals, thus, bear relevance only for the purpose of changing the model output and would be irresponsible to apply to the target itself. For example, a medical condition manifesting in a certain symptom (e.g. a genetic condition affecting behavior) does not imply that the treatment of the symptom will also affect the condition even though the value of the treatment variable affects the model output.}
Finally, a prerequisite for identifying confounding variables causally influencing both in- and outputs of a model is to be able to recognize features with a statistical association to the target in the first place. The inability of suppressor attributors to distinguish such features from suppressor variables, as discussed here, thus implies that XAI methods cannot answer causal questions, such as questions related to confounding.

In both examples, any intervention on $X_1$, $X_2$, or $H$ would have no effect on $Y$ in the real world. In Example~B, interventions on $X_2$ would not even affect the model output, as the model is invariant to changes in $X_2$ by construction. In Example~A, possible interventions on $X_2$ through $H$ could affect the model output; however, not in ways that would correlate with changes in $Y$.

\begin{table}[htbp]
\caption{Summary of the results of Kindermans et al.~\cite{kindermansLearningHowExplain2017} and Wilming et al.~\cite{wilming2023theoretical}. Various popular feature attribution methods systematically attribute non-zero importance to suppressor variables that have no statistical association to the target variable. For Shapley values, this property may depend on the chosen value function.}
\vspace{2mm}
\begin{tabular}{l}
\toprule
XAI methods attributing non-zero importance to suppressors\\
\midrule
Shapley Value \citep{shapley1953value} \\
Permutation Feature Importance \citep{breiman2001random} \\
Partial Dependency Plot \citep{hastie2009elements} \\
Gradient \citep{baehrensHowExplainIndividual2010} \\
Faithfulness \citep[Pixel Flipping,][]{samekEvaluatingVisualizationWhat2015} \\
LIME \citep{ribeiroWhyShouldTrust2016} \\
SHAP \citep[Marginal Expectation,][]{lundbergUnifiedApproachInterpreting2017} \\
Counterfactuals \citep{wachterCounterfactualExplanationsOpening2017} \\
Integrated Gradient \citep{sundararajanAxiomaticAttributionDeep2017} \\
LRP/DTD \citep{bachPixelWiseExplanationsNonLinear2015,montavonExplainingNonLinearClassification2017} \\
SHAP \citep[Conditional Expectation,][]{aasExplainingIndividualPredictions2020}\\
\bottomrule
\end{tabular}
\label{tab:tab1}
\end{table}

\section*{Structural limitations of current XAI research}
\revif{The SAP presents an example of a formal criterion (here, a necessary condition) for XAI fitness for a variety of downstream purposes. Through joint theoretical analyses of data-generating processes, ML models, and feature attribution methods as well as through simulations using synthetic data with known ground-truth explanations, it was further demonstrated that the SAP presents a testable property that enables judgments of the correctness of specific methods for these purposes.}
These techniques are, however, not currently part of the standard toolkit for assessing the quality of explanations and XAI methods, pointing to the following fundamental structural limitations in the way the field assesses itself. 

\paragraph*{Lack of formal problem definitions}
The current XAI terminology uses the term ``explanation'' indiscriminately in different contexts. 
This lack of differentiation gives rise to equivocality of evaluation frameworks and is reflective of a deeper {absence of well-defined problems for XAI to solve.} Even though XAI methods are frequently proposed to serve purposes such as those listed in \nameref{sec:purposes}, it is rarely stated what concrete types of conclusions can be drawn from the explanations provided by any particular method, and under which assumptions each conclusion is valid.
Instead, various popular XAI methods are purely algorithmically defined without reference to a formal problem or a cost function to be minimized, leading to circularity where the method defines the problem it solves.
In their work, Ribeiro et al.~\cite{ribeiroWhyShouldTrust2016} do not define what the correct features for LIME to highlight would be -- the algorithm itself is considered to be the definition of feature importance.

\paragraph*{Existing theory spares out notions of explanation correctness} 
Existing theoretical work has postulated axioms that are desirable for XAI methods to fulfill. For example, according to Sundararajan et al.~\cite{sundararajanAxiomaticAttributionDeep2017}, a method satisfies sensitivity, if a) for every input and baseline that differ in one feature but have different predictions, the differing feature is given non-zero importance, and if also b) the importance of a variable is always zero if the function implemented by the deep network does not depend (mathematically) on it. 
Axioms like this encode meaningful sanity checks but do not provide a notion of correctness or utility-for-purpose of an explanation. 
Several authors have proposed to close this gap by describing criteria for the ``faithfulness'' or ``fidelity'' of XAI methods. These concepts, however, are often not formulated in mathematically stringent form \citep[see,][]{guidottiSurveyMethodsExplaining2019a,jacoviFaithfullyInterpretableNLP2020}. Moreover, faithfulness is insufficient to serve the purposes mentioned in \nameref{sec:purposes}, as we note further below.


\paragraph*{XAI methods ignore data distribution and causal structure}
With few exceptions, XAI methods are applied post-hoc to model weights or outputs only. However, a model's behavior cannot be meaningfully interpreted without knowledge of the correlation or causal structure of its training data \citep{haufe2014interpretation,weichwald2015causal,karimi2021algorithmic,wilming2023theoretical}. The same model weights that cancel out \revi{target-irrelevant noise} in Examples~A and B (see \nameref{sec:examples}) would have a completely different interpretation when applied to features \revi{that are mutually statistically independent}, where their role would be to aggregate independent pieces of target-related information.

Most XAI methods explicitly or implicitly assume statistically independent features. This is in line with the common conception that the main mechanism by which multivariate models achieve their predictive power is to combine (independent) information in order to leverage non-linear interactions in the data. However, this perspective overlooks that an equally important task of multivariate models is to denoise interrelated features, which is achieved by \emph{removing} task-irrelevant signals. 
Incorrectly assuming independence can lead to violations of the SAP, and, thereby, to all of the described misinterpretations.

\paragraph*{``Interpretable'' models share limitations of XAI}
Various authors \citep[e.g.,][]{caruana2015intelligible,rudinStopExplainingBlack2019} make a distinction between ``explainable AI'', which would include post-hoc feature attribution methods, and ``interpretable AI'', which would include model architectures that are  presumed to be intrinsically understandable to humans due to their simplicity. The latter are also occasionally referred to as ``glassbox'' models \citep{rai2020explainable}, and examples include linear models, \revi{generalized additive models (GAMs)}, models with sparse coefficients, and decision trees. However, what concrete interpretations such models are thought to afford is rarely stated. In the above Examples A and B, the Bayes-optimal linear models are uniquely defined and assign non-zero weights to suppressor variables, prohibiting certain desired interpretations and precluding certain actionable consequences, \revi{as demonstrated in Haufe et al.~\cite{haufe2014interpretation} and Wilming et al.~\cite{wilming2023theoretical}. Analogous fallacies apply to GAMs as shown by Clark et al.~\cite{clarkcorrecting2025}. These works highlight that trained ML models, no matter how simple their structure, cannot be univocally interpreted without knowledge of the causal structure of their training data. The standard interpretation of models such as linear models and GAMs implicitly assumes statistically independent features, thereby sharing a fundamental limitation with post-hoc feature attribution methods.}

Given these challenges, an often assumed ``tradeoff'' between predictiveness and ``interpretability'' of models \citep{shmueliExplainPredict2010,del_giudice_prediction-explanation_2024} appears to be misleading. Rather, one needs to acknowledge that even simple models cannot be  unambiguously interpreted without knowledge of the distribution or underlying data generating process of their training data. This is not to say, though, that simple models cannot easen certain interpretations. For example, sparse models can significantly reduce the number of features, the behavior of which, needs to be investigated \citep{doshi2017towards}. Notwithstanding, sparsity alone does not guarantee that a feature or neuron with non-zero weight is not a suppressor \citep{haufe2014interpretation}.

\paragraph*{Empirical evaluation frameworks spare out explanation correctness}
Existing frameworks for empirical XAI evaluation \citep[e.g.,][]{hedstrom2023quantus} often primarily focus on secondary desiderata such as robustness of explanations instead of providing quantifiable notions of correctness. Nevertheless, ``faithfulness'' metrics are widely considered to be suitable surrogates for assessing explanation correctness.
The most widely adopted operationalization of faithfulness is that the ablation (e.g., omission or obfuscation) of an important feature will lead to a drop in a model's prediction performance. The presence of such a drop is then used to assess ``correctness''. Popular perturbation approaches include permutation feature importance \citep{breiman2001random}, stability selection \citep{meinshausen2010stability}, pixel flipping \citep{samekEvaluatingVisualizationWhat2015}, RemOve And Retrain \citep[ROAR,][]{hookerBenchmarkInterpretabilityMethods2019c}, and Remove and Debias \citep[ROAD,][]{rong2022consistent}, and prediction difference analysis \cite[e.g.,][]{blucher2022preddiff}. A variation is the model parameter randomization test \citep[MPRT,][]{adebayoSanityChecksSaliency2018}. 

Despite the simplicity and intuitive appeal of faithfulness metrics, Wilming et al.~\cite{wilming2023theoretical} show that removal or manipulation of $X_2$ in Examples~A and B leads to an inevitable decrease in classification performance, which would lead XAI methods attributing high importance to $X_2$ to appear as faithful. This is because current faithfulness metrics have limited ability to take the data-generating process and the resulting dependency structure in the data properly into account. In that respect, XAI methods and the metrics used to assess their performance share identical limitations. In fact, the idea of ablation is also central to certain feature attribution methods such as Shapley values.

\paragraph*{Insufficiency of real data to validate XAI}
Real datasets are often used for empirical evaluations of XAI methods. In such studies, no ground-truth for the inherently unsupervised XAI problem is available for which reason faithfulness metrics (see above) or human judgement (see below) is used, possibly leading to \revi{biased evaluations and incorrect assessments of explanation correctness}.

\paragraph*{Insufficiency of human judgment to validate XAI}
Several studies \revi{\citep[e.g.,][]{doshi2017towards,
holzingerCausabilityExplainabilityArtificial2019,biessmann2021quality}} consider human judgment for XAI validation, where human experts either annotate inputs ex ante to provide ground-truth explanations or are asked to judge the quality of explanations ex post. While important, such approaches are insufficient as (sole) validations due to the possibility of both Type-I and Type-II errors in human judgments. For example, there may be complex statistical patterns in the data that are leveraged by ML models but {are} (currently) unknown to humans. \revi{This may lead an expert to reject a correct explanation.}  
Human-computer interaction studies are considered an objective way to quantify the added value of AI explanations by some authors \cite[e.g.,][]{jesus2021can}. Such studies compare the joint performance of a human user with access to an XAI with the performance of the user knowing only the outcome of the AI's prediction, the performance of the user alone, and the performance of the AI alone. \revi{However, there are a growing number of studies reporting no correlation between the presence of explanations and combined human-XAI task performance, no correlation between explanation-based human prediction of AI performance and actual AI performance, and no correlation between explanation-induced human trust in AI decisions and actual AI performance \citep{buccinca2020proxy, bansal2021does}. These results speak to the presence of a variety of human biases and psychological factors that hamper attempts to objectively evaluate XAI methods using human judgment alone. In fact, Bansal et al.~\cite{bansal2021does} find that `humans will accept the AI’s recommendation, regardless of its correctness', while Trout \cite{trout2002scientific} discusses how human cognitive biases can generally lead to a wrong sense of understanding incorrect explanations. Notably, overconfidence in XAI explanations can lead to circular reinforcement of wrong beliefs, whereby humans may adapt their judgment to incorrect explanations, ultimately harming scientific knowledge discovery and theory building. As an example, consider a model using an uninformative suppressor feature to remove non-discriminative variance from a target-informative feature. Since this suppression relationship is stable, an XAI method may consistently highlight the suppressor as being important for the prediction. Without further information about the role of the suppressor in the model, this may lead the receiver of an explanation to erroneously conclude that the suppressor carries indispensible discriminative information.}

\paragraph*{Algorithm-first development}
A common paradigm of XAI development is to start with the design of an algorithm and then to demonstrate its utility for various purposes by applying it to selected datasets and models. 
This approach opens the door to experimenter biases due to implicit subjectivity in the choice of the experiments performed and reported. Thereby it becomes possible that capabilities attributed to XAI methods are not systematic but coincidental. 

\paragraph*{High-level nature of existing ML testing and certification frameworks}
Existing efforts to establish processes for the development of trustworthy XAI such as the artificial intelligence assessment methods (DAISAM) guidelines \citep{oala2021machine} established by WHO and ITU, a pre-standard of the German Institute for Standardization (DIN) on explainability \citep{DIN_SPEC_92001-3:2023-04}, explainability fact sheets \citep{sokol2020explainability}, and the Z-inspection framework \citep{amann2022explain,vetter2023lessons} remain on a relatively abstract level and do not provide concrete rules for the proper use case-specific deployment of XAI in practice.



\section*{Towards using XAI for well-defined purposes}
Besides suppressor attribution, XAI methods have been criticized in many further ways~\citep[e.g.,][]{ghassemi2021false, sokolOneExplanationDoes2020a, weberXAITrouble2024,freiesleben2023dear,afroogh2026beyond}. 
For example, the low robustness and consistency of XAI explanations has been noted \citep{babic2021beware}. 
Moreover, explanations provided by different XAI methods are often found to be inconsistent. This can be used by an adversary (e.g., the provider of an ML algorithm in need to explain a decision to a user) to provide arbitrary explanations \citep{bordt2022post}. Similarly, a wealth of quality metrics is available to measure properties such as faithfulness \revi{which are observed to be inconsistent in their ranking of XAI methods \cite{hedstrom2023meta}.} It has been noted that developers of XAI methods could present their own method as being particularly faithful by optimizing the choice of metric \citep{bluecher2024decoupling}.
It has also been pointed out that XAI methods can be manipulated to yield arbitrary explanations \citep{dombrowski2019explanations,xin2024you}. In image prediction tasks, XAI explanations are frequently observed to resemble results of simple edge detection filters \citep[e.g.,][]{adebayoSanityChecksSaliency2018,kauffmann2022clustering, clark2024xai,clark2026feature}. As the results of a controlled study, \citep{clark2026feature} report a further bias towards visually salient image features. The authors note that XAI methods predominantly attribute importance to certain image features regardless of whether these features provide target-related information, represent nuisances unrelated to the target, or were entirely absent during model training.
Many XAI methods also come in multiple variants, and the criteria for choosing methods and their hyperparameters are often not well justified or documented. 

The fundamental limitation of the field, though, is the lack of formal specifications of XAI problems. 
To ensure the fitness of XAI methods for their intended purposes by design,
we argue that the current paradigm of algorithm-first development should be \revi{superseded by a requirement-driven XAI development and validation process \citep[see also][]{DIN_SPEC_92001-3:2023-04}.}
Such a process might consist of the following six steps : 
\begin{enumerate}
    \item Assessing the use case-specific information needs of users and stakeholders. 
    \item Defining the formal requirements and the XAI problems that address these information needs. 
    \item Designing suitable methods to solve these concrete XAI problems.
    \item Performing theoretical analyses, adhering to the formal requirements.
    \item Performing empirical validation using appropriate ground-truth benchmarks.
    \item Improving the methods concerning further desiderata \revi{such as robustness}. 
\end{enumerate}

\noindent \revi{While such a systematic process needs to be carefully developed and refined in a community-wide effort, the remainder of this section provides preliminary considerations on the implementation of its constituents, presents relevant prior work and examples that could be considered as successful partial implementations of individual steps, and discusses challenges and possible limitations of XAI formalization.}

\paragraph*{\revi{Assessing stakeholders' information needs}}
It is unreasonable to call a mapping from input features to real numbers an explanation without endowing these numbers with a well-defined formal interpretation \citep[e.g.][]{murdochDefinitionsMethodsApplications2019a}. 
As indicated above, different stakeholders, such as ML developers, users (e.g., physicians or patients), and regulators, may intend to use XAI for different purposes associated with different information needs. These needs may concern properties of a given ML model, its training data, a given test input, or combinations of these, and may differ between use cases. 

\revi{For example, to perform model diagnostics and quality control, a regulator may want to assess whether a protected attribute such as sex or race unduly affects model decisions in a hiring context. For a similar purpose, a physician may want to make sure that a clinical prediction model does not rely on confounded features lacking biological relevance. An ML developer, on the other hand, may be primarily interested in identifying the set of all features actually used by the model for the purpose of pruning unused features from the training data and model.}

\revi{No single explanation can be the answer to all three questions, and no current XAI method can serve all three purposes at once. Most existing feature attribution methods aim to identify the set of features actually used by a model, however, neither addressing what specific role any given feature plays in the underlying data-generating process, nor \emph{how} and \emph{why} a given model uses that feature. If a model used in hiring puts a non-zero weight on a protected attribute, it still needs to be clarified whether that attribute indeed contains information about the target (e.g. work performance) that is exploited by the model, or whether that attribute rather carries a target-irrelevant signal that the model extracts in order to remove it from its output, effectively to make the model invariant to that attribute. Likewise, if a model is found to rely on a feature suspected to be confounded, it needs to be clarified whether that feature is indeed confounded or actually a genuine cause or effect of the target, or even a suppressor. Whether a variable has any of these properties is determined by the data generating process, which describes the causal relations between variables. In turn, these causal data properties determine whether, how and why prediction models use each variable.}

\revi{Thus, in addition to questions about the model function $f_{\bm \theta}$ itself, which are targeted by classical attribution methods, stakeholders may require extensive information about additional properties of the data and the way model and data interact. XAI developers need to assess these stakeholder needs using, for example, interviews, questionnaires, and inter-disciplinary panels. Ultimately, XAI methods not systematically addressing common information needs will be of little value with respect to specific explanation goals, and for ML quality control in general.}

\paragraph*{Formalizing XAI \revi{problems}}
\revi{To enable the targeted development of XAI methods tailored to specific purposes, informal information needs communicated by stakeholders need to be translated into formal specifications and requirements, which will inevitably lead to distinct XAI problems. In that sense, ``explanation'' is understood here as an umbrella term describing the provision of information to stakeholders. We acknowledge that there can be multiple distinct notions of explanation, and thus explanation correctness, depending on the information requested by stakeholders and provided by XAI. We note that, as the correctness of a formalization cannot itself be readily formally validated, it is critical to employ alternatives measures to ensure that it matches stakeholder intentions. To this end, the assessment of stakeholder needs and their subsequent formalization should go hand in hand and is best carried out by seeking consensus in inter-disciplinary and inter-professional teams  \citep[c.f.,][]{zicari2021z}.}

\revi{An example for XAI problem formalization is the work of Karimi et al.~\cite{karimi2021algorithmic} on minimal algorithmic recourse, defined as the identification of minimal interventions in model inputs that lead to desired changes in model output. The authors formulate this problem as a constrained optimization over interventions in a given structural causal model (SCM). 
It is worth noting, though, that counterfactual and recourse explanations may suggest interventions on causal effects of the target or completely non-informative suppressor variables, changes in both of which would have no effect on the target in the real world \citep{wilming2023theoretical}. More generally, previous work\cite{wilming2022scrutinizing,wilming2024gecobench,borgonovoExplainingClassifiersMeasures2023} proposes the SAP as a necessary, though not sufficient, requirement for important features in the context of the explanation purposes discussed here.}


\revi{In analogy to these examples, future work will develop formal specifications for a broader variety of XAI problems, each addressing different stakeholder needs. 
Importantly, such formalizations can also generate theoretical insight about the identifiability of the desired information. Karimi et al.~\cite{karimi2020algorithmic} show that algorithmic recourse in general requires perfect knowledge of the data-generating SCM, which is unidentifiable from purely observational data and rarely known in practice. Based on this insight, the authors derive algorithms that are applicable under relaxed assumptions such as partial knowledge of the SCM.}


\paragraph*{\revi{Development and theoretical analysis of XAI algorithms}}
\revi{Given formal problem or requirement specifications, it becomes possible to theoretically analyze existing XAI algorithms with respect to established formal criteria. This has led to the identification of various undesired properties and systematic failure modes of existing XAI methods \citep[e.g.,][]{sixtWhenExplanationsLie2020b, bilodeauImpossibilityTheoremsFeature2024a}. 
Kindermans et al.~\cite{kindermansLearningHowExplain2017}, Wilming et al.~\cite{wilming2023theoretical}m, Frye et al.~\cite{fryeAsymmetricShapleyValues2020}, and Martin and Haufe~\cite{martin2026cc} analyzed popular feature attribution methods and found that many do not fulfill the SAP property in general. Such analyses can help to identify theoretical shortcomings and guide the development of novel, improved methods.}

\revi{Similarly, it may be possible to devise algorithms that meet formal criteria. The Pattern approach \citep{haufe2014interpretation} relates a fitted linear model univariately to each individual input feature, thereby avoiding possible misinterpretations due to correlated features. Pattern can consequently be shown to correctly reject suppressor variables in the studied Examples A and B. Recent generalizations such as PatternGAM \citep{clarkcorrecting2025} and PatternLocal \citep{gyolbeminimizing2025} extend the Pattern concept to non-linear models and have shown performance gains in empirical benchmarks involving non-linear data \citep{clark2024xai}. These feature attribution methods fulfill the SAP under well-defined conditions, thereby ensuring that features with high attribution represent sensible starting points, if not solutions, for a variety of popular explanation goals.}

\paragraph*{XAI benchmarking using ground-truth data}
\revi{While formal problem specifications are indispensable to establish XAI as an exact science, formal verification of algorithmic solutions may sometimes be infeasible or considered insufficient to assess a tool's practical utility (see \citep{oberkampf2010verification,imbert2023formal} for discussions of formal verification methods in computer science). In such cases, concordance with formal requirements may be assessed empirically.}
It is often possible to design ground-truth data that share realistic aspects of observational data yet are generated from controlled parametric distribution such that the correct explanation is partially or fully determined by statistical properties of the data.
Various authors propose datasets in which the features having a statistical association with the target are known by construction \citep{ismailInputCellAttentionReduces2019c, hookerBenchmarkInterpretabilityMethods2019c,yalcin2021evaluating,wilming2022scrutinizing,arras2022clevr,zhou2022feature,clark2024xai,budding2021evaluating,oliveira2024benchmarking, oramas2018visual}. This can be used to empirically assess explanation correctness with respect to the SAP, and to quantify explanation performance. Fok and Weld
\cite{fok2023search} construct prediction problems with corresponding textual and visual explanations that can be verified by the user. 
Oramas et al.~\cite{oramas2018visual} introduce synthetic image datasets, where color manipulations of predefined object parts serve as class-related features defining ground-truth explanations.
In \cite{wilming2022scrutinizing} and \cite{clark2024xai}, a range of popular XAI methods in combination with distinct neural network architectures were benchmarked on linear and non-linear image classification problems. In \cite{oliveira2024benchmarking}, structural magnetic resonance imaging (MRI) data were superimposed with synthetic brain lesions and the effect of pre-training on explanation performance in lesions classification tasks was studied. Wilming et al.~\cite{wilming2024gecobench} introduce a gender-balanced text dataset and associated gender classification tasks, which allows for quantifying explanation performance and biases in explanations. These datasets are publicly available. 

\revi{Note, though, that empirical benchmarking using ground-truth data can only provide partial validation due to the impossibility to cover all realistic aspects and configurations of real-world settings. It should therefore be complemented by theoretical analyses whenever possible. Notwithstanding, benchmarks can be very useful to identify failure modes and invalidate certain approaches through counterexamples.}

\paragraph*{\revi{Improving secondary XAI quality indicators}}
Methods that are theoretically and/or empirically validated with respect to given XAI problems or goals should also be theoretically analyzed, quantitatively benchmarked, and algorithmically improved with respect to secondary quality indicators such as robustness, fairness, uncertainty calibration, and alignment with human knowledge in real-world settings.
To this end, dedicated benchmarking frameworks such as Quantus \citep{hedstrom2023quantus} can be of use. Moreover, it is crucial to present explanations in ways that are aligned with human cognition and social norms \citep{miller2019explanation}. \revif{Note that, although these goals do not necessarily contradict each other, it is an open question whether they can be jointly accomplished in general.}. 
Finally, it can be worthwhile to expand the range of applicability of validated XAI methods. Along these lines, recent work has extended the concept of activation patterns \citep{haufe2014interpretation} to non-linear and local explanation domains \citep{clarkcorrecting2025,gyolbeminimizing2025}.

\section*{Discussion and Outlook}

Just as ML in general, the field of XAI is fast-paced with clever novel methodological developments and empirical validation approaches being introduced each year. Recent advancements in confounder detection \citep{janzing2018detecting2}, generative modeling \citep{hvilshoj2021ecinn,}, and causal representation learning \citep{scholkopf2021toward,ahuja2023interventional} promise to address some of the limitations presented here.
The systematic formalization and scrutinization of the field of XAI is a wider effort that will eventually make it possible to objectively assess the ability of approaches to solve specific XAI problems. This may lead to XAI-based workflows that can indeed be used to systematically perform quality assurance for ML -- and that may eventually find their way into ML production processes and industry standards.

Theoretical and empirical analyses of simple data-generating models have shown that popular feature attribution methods can systematically fail to answer important questions about data and ML models. The main technical limitation of existing feature attribution methods is the explicit or implicit assumption of feature independence, causing false interpretations in the considered examples.
On a more general level, the field of XAI is impeded by the current paradigm of algorithm- instead of problem-driven development and the lack of formal notions of explanation correctness. These limitations are not specific to feature attribution methods but are shared by other XAI paradigms such as concept- or example-based explanations.
Researchers should formally define the specific problems that XAI should solve and design methods accordingly. Synthetic data with ground-truth explanations can play an important role in (in)validating XAI methods.

\section*{Acknowledgments}
This work was supported by the European Research Council (ERC) under the European Union's Horizon 2020 research and innovation programme (Grant agreement No. 758985), the European Partnership on Metrology (Grant agreement 22HLT05), and the Metrology for Artificial Intelligence in Medicine (M4AIM) programme funded by the Federal Ministry for Economy and Climate Action (BMWK) in the frame of the QI-Digital Initiative. 

\section*{Author Contributions}
All authors wrote and approved the manuscript. RW prepared Figure~\ref{fig:fig1}.

\section*{Competing Interests}
The authors declare no competing financial or non-financial interests.

\bibliography{used}
\bibliographystyle{naturemag}

\appendix

\section*{Figure legend}

Figure 1: a/b) Data sampled from the generative model (Example A) introduced in \nameref{sec:examples} \citep{wilming2023theoretical} for two different correlations $c$ and constant variances $s_1^2 = 0.8$ and $s_2^2=0.5$. Boundaries of {the} Bayes-optimal decisions are shown as well. The marginal sample distributions illustrate that feature $X_2$ does not carry any class-related information. c) Causal structure of the data in Examples~A (left) and B (right). $X_2$ is a so-called suppressor variable that has no statistical association with the target $Y$, although both influence feature $X_1$, which is called a collider. Figure partially adopted from Wilming et al.~\cite{wilming2023theoretical}.

\section*{Table legend}

Table 1: Summary of the results of Kindermans et al.~\cite{kindermansLearningHowExplain2017} and Wilming et al.~\cite{wilming2023theoretical}. Various popular feature attribution methods systematically attribute non-zero importance to suppressor variables that have no statistical association to the target variable. For Shapley values, this property may depend on the chosen value function.




\end{document}